\documentclass[letterpaper]{article} 
\usepackage{aaai25}  
\usepackage{times}  
\usepackage{helvet}  
\usepackage{courier}  
\usepackage[hyphens]{url}  
\usepackage{graphicx} 
\usepackage{natbib}  
\usepackage{caption} 
\usepackage{algorithm}
\usepackage{algorithmic}

\usepackage{amsmath}
\usepackage{amssymb}
\DeclareMathAlphabet\mathbfcal{OMS}{cmsy}{b}{n}
\usepackage{booktabs} 
\usepackage{xcolor}
\usepackage{multirow}
\usepackage{pifont}
\usepackage{newfloat}
\usepackage{listings}
\DeclareCaptionStyle{ruled}{labelfont=normalfont,labelsep=colon,strut=off} 
\floatstyle{ruled}
\newfloat{listing}{tb}{lst}{}
\floatname{listing}{Listing}
\title{NEST: A Neuromodulated Small-world Hypergraph Trajectory Prediction Model for Autonomous Driving}
\author{
    Chengyue Wang\textsuperscript{\rm 1}\equalcontrib,
    Haicheng Liao\textsuperscript{\rm 1}\equalcontrib,
    Bonan Wang\textsuperscript{\rm 1},
    Yanchen Guan\textsuperscript{\rm 1},
    Bin Rao\textsuperscript{\rm 1},
    Ziyuan Pu\textsuperscript{\rm 2},
    Zhiyong Cui\textsuperscript{\rm 3},
    Chengzhong Xu\textsuperscript{\rm 1},
    Zhenning Li\textsuperscript{\rm 1}\thanks{Corresponding Author}
}
\affiliations{
    \textsuperscript{\rm 1}University of Macau\\
    \textsuperscript{\rm 2}Southeast University\\
    \textsuperscript{\rm 3}Beihang University\\

    chengyue.wang@connect.um.edu.mo, \{yc27979, mc35002, yc37976, binrao\}@um.edu.mo, ziyuanpu@seu.edu.cn, zhiyongc@buaa.edu.cn, \{czxu, zhenningli\}@um.edu.mo
}

\usepackage{bibentry}

\begin{document}

\maketitle

\begin{abstract}
Accurate trajectory prediction is essential for the safety and efficiency of autonomous driving. Traditional models often struggle with real-time processing, capturing non-linearity and uncertainty in traffic environments, efficiency in dense traffic, and modeling temporal dynamics of interactions. We introduce NEST (Neuromodulated Small-world Hypergraph Trajectory Prediction), a novel framework that integrates Small-world Networks and hypergraphs for superior interaction modeling and prediction accuracy. This integration enables the capture of both local and extended vehicle interactions, while the Neuromodulator component adapts dynamically to changing traffic conditions. We validate the NEST model on several real-world datasets, including nuScenes, MoCAD, and HighD. The results consistently demonstrate that NEST outperforms existing methods in various traffic scenarios, showcasing its exceptional generalization capability, efficiency, and temporal foresight. Our comprehensive evaluation illustrates that NEST significantly improves the reliability and operational efficiency of autonomous driving systems, making it a robust solution for trajectory prediction in complex traffic environments.
\end{abstract}

%

\section{Introduction}

\begin{figure}[t]
        \centering
	\includegraphics[width=0.9\linewidth]{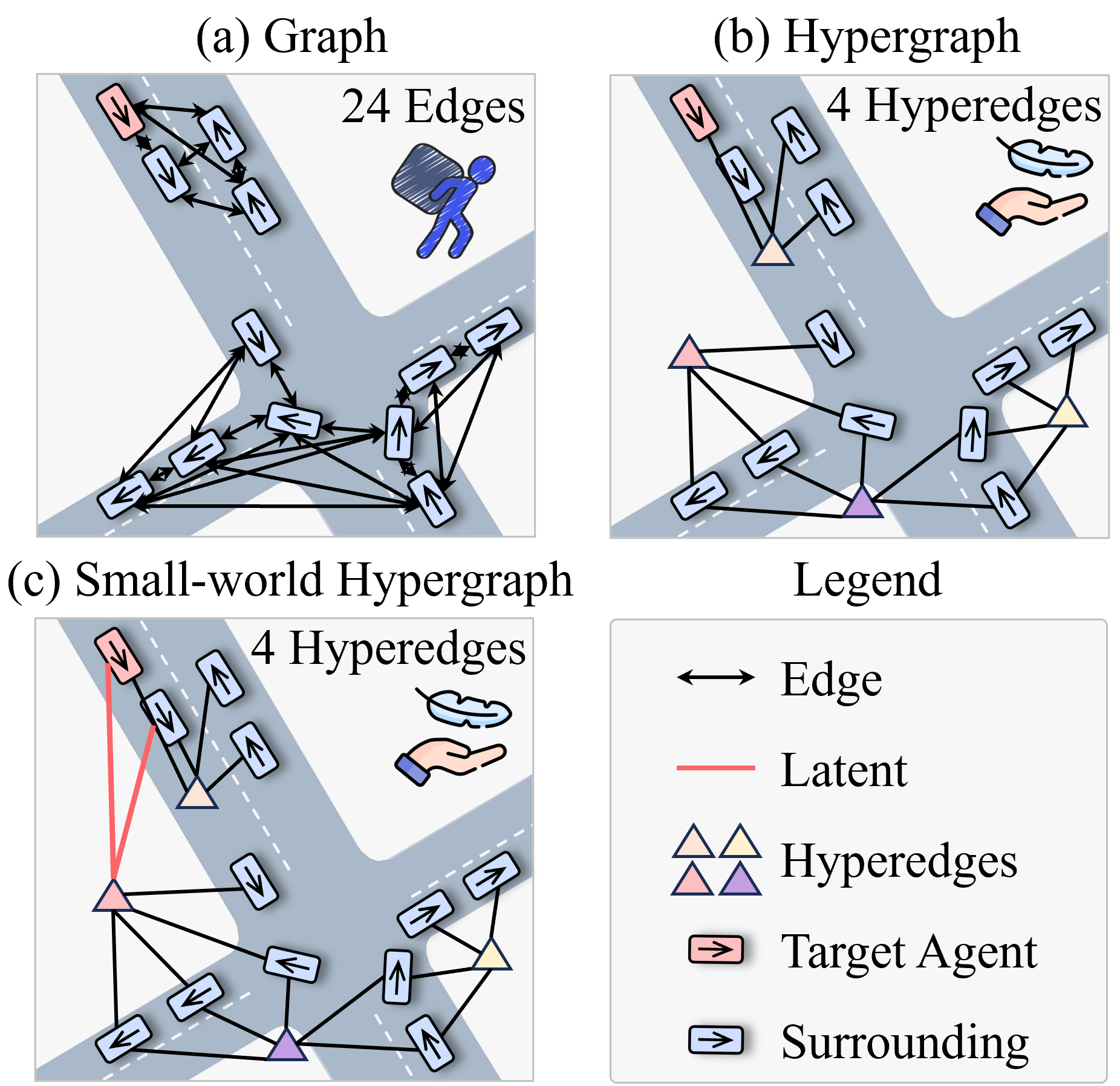}
	\caption{Comparison of interaction modeling methods. (a) Traditional graph-based approach requires 24 edges to represent agent interactions. (b) Hypergraph method reduces this complexity to 4 hyperedges by grouping interactions. (c) Our Small-world Hypergraph further refines this by incorporating latent connections, effectively capturing long-range interactions typical of traffic scenarios. The use of hyperedges and latent links allows for a more efficient and comprehensive representation of agent interactions.}
        \label{toutu}
\end{figure}

The ability to predict the future paths of surrounding vehicles with precision is not just an academic pursuit—it's a life-saving technology at the heart of autonomous driving (AD). In the bustling dance of traffic, where vehicles, pedestrians, and cyclists move in a complex choreography, even a moment's hesitation can lead to disaster \cite{pourkeshavarz2024adversarial}. Thus, the quest for an efficient model that can foresee these movements with high fidelity is more than a goal; it's a necessity for the next leap in vehicular autonomy \cite{duan2022complementary}.

Traditional trajectory prediction models have paved the way, yet they falter under the unpredictable and multifaceted nature of real-world traffic \cite{wong2024socialcircle}. One of the significant limitations of current trajectory prediction models is their inability to effectively capture the temporal dynamics of interactions \cite{chen2022intention}. Traffic behavior is inherently dynamic, with interactions evolving over time in complex, often unpredictable ways. For instance, a vehicle's maneuver can trigger a cascade of reactions from surrounding vehicles, each adjusting its trajectory based on the observed behavior. Traditional models \cite{letter2017efficient,liu2020optimizing}, however, often rely on static snapshots of traffic, failing to account for how these interactions unfold over time. This leads to suboptimal predictions that do not accurately reflect the real-world progression of traffic scenarios, ultimately compromising the safety and efficiency of AD systems.

Moreover, these models frequently struggle with the inherent non-linearity and uncertainty of traffic environments \cite{wang2022socialsurvey}. Traffic dynamics are influenced by a myriad of variables, including sudden stops, erratic driving patterns, and the unpredictable behavior of pedestrians and cyclists. The non-linear nature of these interactions introduces chaos that many current models cannot adequately address. For example, a pedestrian suddenly crossing the road or a cyclist weaving through traffic can dramatically alter the flow of vehicles, creating a highly unpredictable environment. Traditional models often oversimplify these interactions, leading to inaccurate predictions during unforeseen events \cite{wong2024socialcircle}.

Efficiency represents another significant challenge, especially in heterogeneous urban environments where traffic density is extremely high. In such scenarios, various heterogeneous traffic agents—vehicles, pedestrians, and cyclists—interact, resulting in an exponential increase in the types of interaction relationships \cite{ramezani2015dynamics}. Each agent may be involved in multiple interaction groups, with each group comprising different numbers of agents. Traditional methods of modeling interactions are often confined to predefined relationships \cite{lv2023ssagcn,xu2023mvhgn}, limiting their efficiency in capturing diverse and dynamic interactions. This underscores the necessity for an efficient paradigm that can accommodate diverse interaction relationships in real time.

To address these challenges, we propose the NEST (\textbf{Ne}uromodulated \textbf{S}mall-world Hypergraph \textbf{T}rajectory Prediction) model for autonomous driving. This framework combines the intricate structure of Small-world Networks \cite{watts1998collective} with the expansive reach of hypergraphs \cite{gao2022hgnn+} to enhance interaction modeling and trajectory prediction. Small-world Networks, characterized by high clustering and short path lengths, effectively capture both local and long-range interactions among vehicles, leading to a more accurate representation of traffic dynamics (as illustrated in Figure \ref{toutu}), where even distant vehicles can influence each other's trajectories.

The Neuromodulator component adapts this structure to reflect the dynamic and diverse nature of real-world traffic \cite{grossman2022neuromodulation}, ensuring that the model remains robust and responsive to changing conditions. Hypergraphs extend beyond traditional graph structures by allowing hyperedges to connect multiple nodes simultaneously, which is crucial for modeling the collective behavior of traffic agents \cite{xu2022groupnet}, such as groups of pedestrians or a convoy of vehicles merging into a lane. By integrating Small-world Networks and hypergraphs, our model not only predicts trajectories but also provides a comprehensive understanding of vehicular interactions. This fusion addresses the limitations of existing models, offering enhanced efficiency to handle dense traffic, temporal foresight to capture evolving interactions, and contextual integration to consider relevant external factors. Moreover, the dynamic adaptability of our model ensures it can generalize beyond its training datasets, making it suitable for a wide range of scenarios.

To sum up, our contributions are threefold:
\begin{itemize}
    \item We introduce a Small-world Network that effectively captures both local and long-range interactions among traffic agents. The Neuromodulator within the NEST model dynamically adjusts the network to contextual information, enhancing adaptability in diverse and dynamic traffic scenarios.
    \item We propose a novel Hypergraph Neural Network (HGNN) for interaction learning. Our unique hyperedge set structure combines hyperedges of different relationship types, providing an efficient framework for modeling complex and diverse interaction dynamics.
    \item We conduct extensive validation of the NEST model using multiple real-world datasets, including nuScenes, MoCAD, and HighD. The results consistently demonstrate that the NEST model outperforms existing methods in various traffic scenarios, showcasing its exceptional generalization capabilities and reliability.
\end{itemize}

\begin{figure*}[t]
        \centering
	\includegraphics[width=0.98\textwidth]{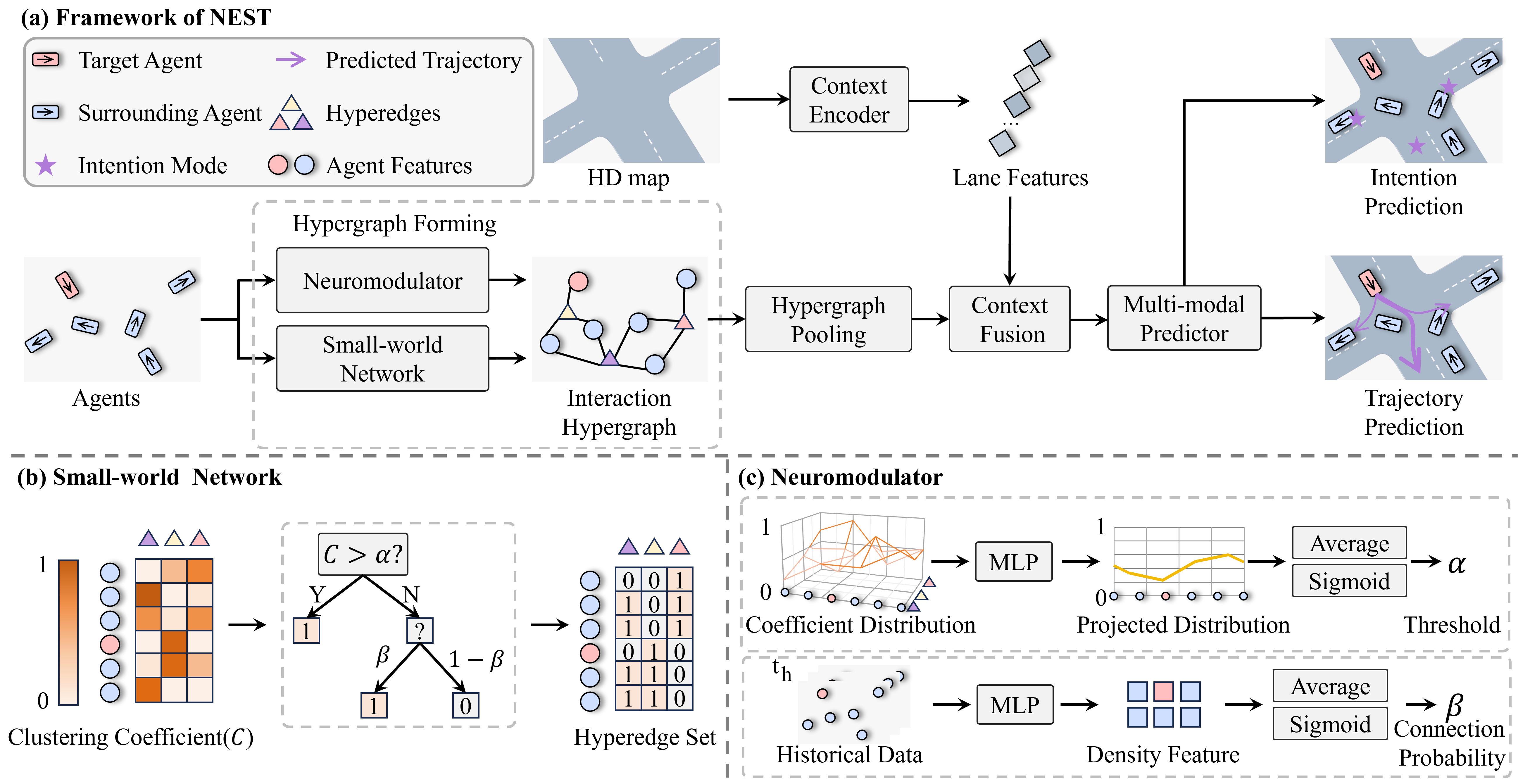}
	\caption{The overview of the proposed NEST model. Panel (a) illustrates the overall framework of NEST, which takes agent historical data and high-definition (HD) maps as inputs. The NEST model processes these inputs with four key modules: Hypergraph Forming, Hypergraph Pooling, Context Fusion, and a Multi-modal Predictor. These modules work in tandem to predict the target agent's intention and future trajectory. Panels (b) and (c) provide a detailed breakdown of the components involved in forming the interaction hypergraph, specifically focusing on the Neuromodulator and the Small-world Network.}
        \label{kuangjiatu}
\end{figure*}
\section{Related Work}\label{Related Work}
With the significant advancements in deep learning technologies across various domains of autonomous driving (AD), it is natural that researchers have increasingly applied these techniques to the field of trajectory prediction. Initially, trajectory prediction was approached as a time-series forecasting problem, leading to the use of neural networks designed for handling sequential data, such as Recurrent Neural Networks (RNNs) and Long Short-Term Memory (LSTM) networks, as encoders and decoders in these models \cite{lan2024hi, liao2024physics}. A pivotal development in this area was the introduction of Social LSTM by Alahi et al. \cite{alahi2016social}, which first brought the concept of social forces into trajectory prediction, highlighting the importance of understanding interactions between agents.

Building on this foundation, researchers shifted their focus towards modeling interactions from both spatial and temporal dimensions \cite{liao2024cognitive, liao2024less, xu2022groupnet, chen2024q}. The use of attention mechanisms further advanced the field, enabling models to capture complex interactions among agents \cite{liao2024mftraj, lan2024hi, liao2024cdstraj}. For instance, the Nettraj model \cite{liang2021nettraj} introduced Spatiotemporal Attention Mechanisms with a sliding context window to capture both short- and long-term interactions. However, these attention-based methods often rely on grid-based approaches \cite{liao2024human, liao2024cognitive2}, segmenting the scene into fixed grids, which limits their applicability to structured road topologies.

To address the limitations of grid-based methods, Graph Neural Networks (GNNs) have emerged as a promising approach, particularly suited for modeling interactions in unstructured and complex urban environments. GNNs excel at representing interactions between vehicles in scenarios where road structures are not strictly defined. For example, the SFEM-GCN model proposed by Du et al. \cite{du2024social} incorporates three types of graph structures to capture semantic, positional, and velocity information of traffic agents, thereby offering a more comprehensive interaction model. Similarly, the MTP-GO model \cite{westny2023mtp} employs a temporal graph to encode traffic scene information and utilizes neural ordinary differential equations for final predictions. The Social Soft Attention Graph Convolution Network by Lv et al. \cite{lv2023ssagcn} addresses the challenge of handling interactions not only between agents but also between agents and their environment.

Despite their advantages, graph-based models have inherent limitations. They often focus on one-to-one relationships, potentially overlooking broader group dynamics and the interconnected nature of traffic behaviors. Moreover, efficiency issues and computational constraints can become significant, especially in dense urban environments with numerous interacting agents. While GNNs provide a more comprehensive representation of interactions compared to other models, they may still fall short in fully capturing the complexity and dynamic interactions inherent in real-world traffic scenarios.

\section{Methodology}\label{Method}
\subsection{Problem Formulation}
The objective is to predict the future trajectories of surrounding traffic agents within a defined perception range over a specific period. Each agent to be predicted is referred to as the \textit{target agent}, while all other agents within the perception range are considered as \textit{surrounding agents}. 

The NEST model takes two inputs: historical data $X$ spanning the past $t_h$ time steps, and a high-definition (HD) map $M$. The historical data $X = [X_0, X_1, \ldots, X_n]$ includes coordinates, velocity, and acceleration for the target agent $X_0$ and the surrounding agents $X_{i \in [1,n]}$. $n$ denotes the number of agents. The HD map $M$ records the road topology. The output is the future trajectory $Y$ of the target agent over the prediction horizon $t_f$. Predicting the future trajectories of traffic agents involves addressing various uncertainties rooted in the diversity of driving intentions and the unpredictability of motion. The NEST model addresses these uncertainties by generating multiple predicted trajectories, $Y = [Y_1, Y_2, \ldots, Y_K]$, where each $Y_i$ corresponds to a distinct intention mode with an associated probability $P_i$. Here, $K$ denotes the number of intention modes. To capture the unpredictable nature of motion within each intention mode, predictions at each time step $t$ in $Y_i$ are represented by parameters of the Laplace distribution: $Y_i^t = [x_i^t, y_i^t, b_{i,x}^t, b_{i,y}^t]$, where $x_i^t$ and $y_i^t$ are the coordinates, and $b_{i,x}^t$ and $b_{i,y}^t$ are scale parameters reflecting unpredictability of motion.

\subsection{Model Overview}
The proposed NEST model is illustrated in Figure \ref{kuangjiatu}. The Hypergraph Forming module forms the interaction hypergraph $\mathbfcal{G}$ with the Neuromodulator and Small-world Network. Then, the Hypergraph Pooling module captures interaction features $\mathcal{F}_i$ based on the interaction hypergraph $\mathbfcal{G}$. Next, the Context Fusion module combines lane features $\mathcal{F}_l$ from HD maps $M$ with the interaction feature $\mathcal{F}_i$, resulting in a consolidated context feature $\mathcal{F}_c$. Finally, the Multi-modal Predictor uses $K$ generators to predict future trajectories for each intention mode.

\subsection{Hypergraph Forming}
The hypergraph is distinguished by its unique structure where a single hyperedge can link multiple vertices. This unique structure makes it particularly well-suited for modeling group-wise interaction relationships among traffic agents. Each hyperedge defines an interaction group, with each vertex connected by the hyperedge representing an agent within that interaction group.

Forming an appropriate interaction hypergraph $\mathbfcal{G}$ is fundamental for efficiently capturing interactions among agents. We define the interaction hypergraph $\mathbfcal{G}=(\mathbfcal{V}, \mathbfcal{E})$, where $\mathbfcal{V}$ is the set of vertex features, and $\mathbfcal{E}$ is the set of hyperedges. To represent traffic agents as vertices, we define the vertex feature set as $\mathbfcal{V} = \mathcal{F}_a \in \mathbb{R}^{(n+1) \times d}$, where the agent features $\mathcal{F}_a$ are obtained through a transformer-based encoder, and $d$ is the hidden dimension of the feature. Specifically, $\mathbfcal{V} =[\mathcal{V}_0, \mathcal{V}_1, ..., \mathcal{V}_n]$, where the feature $\mathcal{V}_i$ of vertex $i$ corresponds to the feature $\mathcal{F}_{a,i}$ of agent $i$, with $\mathcal{F}_{a,0}$ corresponding to the target agent.

\subsubsection{Small-world Network} 
In traffic scenarios, agents can be influenced not only by nearby agents but also by distant ones through interactions mediated by intermediate agents. For example, in a car-following scenario, the braking of a leading vehicle can trigger a chain reaction among following vehicles. This reflects the small-world property of traffic, where agents are connected through short interaction chains, similar to social networks. A Small-world Network, characterized by this property, is therefore ideal for modeling interaction relationships among traffic agents.

To robustly capture these dynamic interactions, we design a Small-world Network module inspired by the Newman–Watts (NW) model \cite{newman1999renormalization} to generate hyperedges. We determine each agent's membership in interaction groups using the clustering coefficient $\mathcal{C}$, which measures the tendency of a vertex to cluster with its neighbors. The clustering coefficient $\mathcal{C}_{i,j}$ for each vertex-hyperedge pair is defined as:
\begin{equation}
\mathcal{C}_{i,j} = 
\begin{cases} 
1 & \text{if } {C}_{i,j} \geq \alpha \\
0 & \text{otherwise}
\end{cases}
\end{equation}
where $\mathcal{C}_{i,j} = 1$ indicates a definite connection between vertex $i$ and hyperedge $j$, while $\mathcal{C}_{i,j} = 0$ suggests a potential connection that is less certain.

The constructed clustering coefficient matrix $\mathcal{C}$ is analogous to the basic regular network in the NW model. To refine the connections between vertices and hyperedges, we re-evaluate the relationships where $\mathcal{C}_{i,j}$ is uncertain, applying a connection probability $\beta$. The final hyperedges are determined as follows:
\begin{equation}
\mathcal{E}_{i,j} = 
\begin{cases} 
1 & \text{if } \mathcal{C}_{i,j} = 1 \ \text{or} \ (\mathcal{C}_{i,j} = 0 \ \text{and} \ \eta \leq \beta) \\
0 & \text{otherwise}
\end{cases}
\end{equation}
where $\mathcal{E}_{i,j} = 1$ indicates that vertex $i$ is connected to hyperedge $j$, and $\eta$ is a randomly distributed value in $[0,1]$.

In summary, the process of generating the hyperedge set can be formulated as follows:
\begin{equation}
\mathbfcal{E} = \Omega(\mathbfcal{V},\alpha,\beta)
\label{eq7}
\end{equation}
where $\Omega$ is the Small-world Network.

\subsubsection{Neuromodulator} 
Human drivers adapt their behavior to ever-changing driving environments based on external stimuli. Inspired by this adaptability, rooted in cellular neuroregulation mechanisms \cite{vecoven2020introducing}, we developed the Neuromodulator to dynamically adjust the Small-world Network, enhancing its responsiveness to traffic conditions. As shown in Equation \ref{eq7}, the hyperedge set is modulated by two parameters: threshold $\alpha$ and connection probability $\beta$, derived from the agent feature $\mathcal{F}_a$.

The threshold $\alpha$ determines which agents are included in an interaction group. To set this threshold, we analyze the distribution of the clustering coefficient matrix $\mathcal{C}$. The threshold neuromodulator $\Pi_\alpha$ projects $\mathcal{C} \in \mathbb{R}^{(n+1) \times s}$ into a threshold feature space of $\mathbb{R}^{(n+1) \times 1}$ using two MLP layers, where $s$ is the predefined number of hyperedges. The final threshold $\alpha \in [0,1]$ is obtained using a sigmoid function:
\begin{equation}
\alpha = \Pi_\alpha(\mathcal{C})
\label{threhold}
\end{equation}
where $\Pi_\alpha$ denotes the threshold neuromodulator.

The connection probability $\beta$ represents the likelihood of involving additional agents in interaction groups, particularly in high-density traffic. The agent feature $\mathcal{F}_a$ encapsulates traffic density, from which the connection probability $\beta \in [0,1]$ is derived using the connection probability neuromodulator $\Pi_\beta$, similar to the threshold neuromodulator:
\begin{equation}
\beta = \Pi_\beta(\mathcal{F}_a)
\label{connection_probability}
\end{equation}
where $\Pi_\beta$ is the connection probability neuromodulator.

\subsection{Hypergraph Pooling}
The interaction hypergraph explicitly defines the traffic agents in each interaction group, with vertices connected by hyperedges. Unlike traditional graphs, which limit information aggregation to pairs of agents, hypergraphs capture interaction features more efficiently through iterative Vertex-to-Hyperedge and Hyperedge-to-Vertex pooling \cite{xu2022groupnet}. This approach provides a more comprehensive representation of interactions in a traffic scene, leading to more accurate trajectory predictions.

\begin{table}[tbp]
\centering
\resizebox{\linewidth}{!}
{
\begin{tabular}{ccccccc}
\toprule
\text{Model} & Venue & $\text{minADE}_{5}$ & $\text{minADE}_1$ & $\text{minFDE}_1$ \\
\midrule
DLow-AF \cite{yuan2020dlow} &    ECCV & 2.11 & - & - \\
Trajectron++ \cite{salzmann2020trajectron++} &     ECCV & 1.88 & - & 9.52 \\
MultiPath \cite{chai2020multipath} &   CoRL & 1.44 &  {3.16} & 7.69 \\
LDS-AF \cite{ma2021likelihoodbased} &    ICCV & 2.06 & - & - \\
AgentFormer \cite{yuan2021agentformer} &   ICCV & 1.97 & - & - \\
LaPred \cite{kim2021lapred} &   CVPR & 1.53 & 3.51 & 8.12 \\
STGM \cite{zhong2022stgm} &  TITS & - & 3.21 & 9.62\\ 
GoHome \cite{gilles2022gohome} &   ICRA &  {1.42} & - &  {6.99} \\
ContextVAE \cite{xu2023context} &   RAL &  1.59 & 3.54 & 8.24 \\
EMSIN \cite{ren2024emsin} &    TFS & 1.77 & 3.56 & - \\
SeFlow \cite{zhang2024seflow} & ECCV  & 1.38 & - & 7.89 \\
 \midrule
\textbf{NEST} & - & \textbf{1.18} & \textbf{2.97} & \textbf{6.87} \\
\bottomrule
\end{tabular}}
\caption{Performance comparison of various models on nuScenes dataset. \textbf{Bold} values represent the best performance in each category. ``-'' denotes the missing value.}
\label{table:performance_nuscene}
\end{table}

\subsubsection{Vertex-to-Hyperedge}
In this process, the hyperedge serves as a platform for information exchange among vertices in the same interaction group. The Vertex-to-Hyperedge step aggregates features from all connected vertices to form a group feature $\mathcal{F}_g$. To align with the requirements for drivers during interactions, $\mathcal{F}_g$ should include three types of information: the personality of each traffic agent, their intentions, and the possibility of these intentions.

Traffic agents in the same group often adjust their behavior based on others, such as drivers yielding to aggressive counterparts. The agent feature $\mathcal{F}_a$ reflects these personalities. We aggregate the features of all vertices connected to the same hyperedge $j$, denoted as $\mathcal{V} = { \mathcal{V}_i \mid \mathcal{E}_{ij} = 1 }$, using a weighted sum. This sum is then processed by a Multi-Layer Perceptron (MLP) based personality encoder $\mathcal{M}_p$ to extract personality information $\mathcal{I}_p$:
\begin{equation}
\mathcal{I}_p=\mathcal{M}_p (\sum_{\mathcal{V}_i \in \mathcal{V}} \lambda_i \mathcal{V}_i)
\end{equation}
where $\lambda_i$ is the learnable weight.

Obtaining the various intentions of agent $i$, along with the degree of willingness to execute each intention, aids the model in predicting future motion. We define the intention information $\mathcal{I}_i = [\mathcal{I}_{i,1}, \mathcal{I}_{i,2}, ..., \mathcal{I}_{i,K}]$, each $\mathcal{I}_{i,j}$ denotes a specific intention. We formulate intention information $\mathcal{I}_i$ as:
\begin{equation}
\mathcal{I}_{i}=\sigma\left(\left(\mathcal{M}_{\mathrm{i}}\left(\mathcal{I}_a\right)+\xi\right) / \tau\right)
\end{equation}
where $\sigma$ is the softmax function, and aggregated information $\mathcal{I}_a = \sum_{\mathcal{V}_i \in \mathcal{E}_j} \lambda_i \mathcal{V}_i$. $\mathcal{M}_{\mathrm{i}}$ denotes the intention encoder. And following \cite{xu2022groupnet}, we introduce Gumbel distribution $\xi$ and temperature parameter $\tau$. 

The degree of willingness for each intention is measured by $\mathcal{I}_w = \mathcal{M}_{\mathrm{w}}(\mathcal{I}_a)$, where $\mathcal{M}_{\mathrm{w}}$ is the willingness encoder. After obtaining personality information $\mathcal{I}_p$, intention information $\mathcal{I}_i$, and willingness information $\mathcal{I}_w$, the group feature $\mathcal{F}_g$ if formulated as following:
\begin{equation}
\mathcal{F}_g=\mathcal{I}_p \sum_{j=1}^K \mathcal{I}_{i, j}\mathcal{I}_{w,j}
\end{equation} 
where $\mathcal{I}_{i, j}$ and $\mathcal{I}_{w,j}$ are the pairs of intention mode $j$ and its associated willingness. $K$ is the number of intentions.

\subsubsection{Hyperedge-to-Vertex} To enable traffic agents to obtain relevant information from the hyperedges, we employ the Hyperedge-to-Vertex approach. This method updates each vertex feature $\mathcal{V}_i$ based on the group features $\mathcal{F}_{g}$ of the connected hyperedges. During this update process, we consider the group features of all hyperedges linked to the vertex $i$, denoted as $\mathcal{E}={ \mathcal{E}_j \mid \mathcal{E}_{ij} = 1 }$. The update process is described as follows:
\begin{equation}
\mathcal{V}_i \leftarrow \mathcal{M}_{v}\left(\left[\mathcal{V}_i, \sum_{\mathcal{E}_j \in \mathcal{E}} \mathcal{F}_{g,j}\right]\right)
\end{equation}
where $\mathcal{F}_{g,j}$ is the group feature for hyperedge $j$, and $\mathcal{M}_{v}$ denotes an MLP-based vertex encoder. After undergoing $H$ iterations of updates, we obtain the refined features for all vertices. To ensure comprehensive consideration of all agents in the scene, we average these updated features to derive the interaction feature $\mathcal{F}_i$.

\subsection{Context Fusion}
The environmental context, especially the topology of the lane network, is essential for accurate trajectory prediction. To incorporate this, we employ a context fusion module that integrates lane features $\mathcal{F}_l$ with interaction features $\mathcal{F}_i$. A Lane Encoder first processes HD maps to generate lane features $\mathcal{F}_l$. Then, an attention module combines these with the interaction features to produce the final context feature $\mathcal{F}_c$, as defined by:
\begin{equation}
\mathcal{F}_c = \textit{Attn}(Q = \mathcal{F}_i, K = \mathcal{F}_l, V=\mathcal{F}_l)
\label{contextfusion}
\end{equation}

This approach allows the model to account for the critical influence of lane topology on agent trajectories, enhancing the accuracy and reliability of trajectory predictions by integrating interaction and contextual information.

\subsection{Multi-modal Predictor}
Driving behavior is inherently uncertain due to both the diversity of intentions at the macro level and the unpredictability of actions at the micro level. To capture this range of behavior, we use a Multi-modal Predictor, where each generator represents a distinct human intention mode. The micro-level unpredictability within each mode is modeled using a Laplace distribution. We employ $K$ generators, each combining the context feature $\mathcal{F}_{c}$ with the agent feature $\mathcal{F}_{a}$ to predict a trajectory $Y_i$, characterized by a Laplace distribution and its associated probability $P_{i}$.

\begin{table}[tbp]
  \centering
  \setlength{\tabcolsep}{3mm}
  \resizebox{0.95\linewidth}{!}{
    \begin{tabular}{cccccc}
    \toprule
    \multirow{2}[3]{*}{Model} & \multicolumn{5}{c}{Prediction Horizon (s)} \\
\cmidrule{2-6}          & 1     & 2     & 3     & 4     & 5 \\
    \midrule
    CS-LSTM \cite{deo2018convolutional} & 1.45  & 1.98  & 2.94  & 3.56  & 4.49  \\
    NLS-LSTM \cite{messaoud2019non} & 0.96  & 1.27  & 2.08  & 2.86  & 3.93\\
     MHA-LSTM \cite{messaoud2021attention} & 1.25  & 1.48  & 2.57  & 3.22  & 4.20  \\
    CF-LSTM \cite{xie2021congestion} & 0.72  & 0.91  & 1.73  & 2.59  & 3.44 \\
    STDAN \cite{chen2022intention} & 0.62  & 0.85  & 1.62  & 2.51  & 3.32  \\
    WSiP \cite{wang2023wsip} & 0.70  & 0.87  & 1.70  & 2.56  & 3.47  \\
    BAT (25\%) \cite{liao2024bat} & 0.65  & 0.99  & 1.89  & 2.81  & 3.58 \\
    BAT \cite{liao2024bat} &  {0.35}  & \textbf{0.74}  &  {1.39}  &  {2.19} & {2.88}\\
 \midrule
   \textbf{NEST} & \textbf{ 0.32 } &  { 0.75 } & \textbf{ 1.27} & \textbf{ 2.01 } & \textbf{ 2.42 } \\
    \bottomrule
    \end{tabular}%
    }
  \caption{Experimental results on MoCAD. Metric: RMSE}
  \label{mocadresult}%
\end{table}%

\begin{table}[tbp]
  \centering
  \setlength{\tabcolsep}{3mm}
   \resizebox{0.95\linewidth}{!}{
    \begin{tabular}{cccccc}
    \bottomrule
    \multirow{2}[3]{*}{Model} & \multicolumn{5}{c}{Prediction Horizon (s)} \\
\cmidrule{2-6}          & 1     & 2     & 3     & 4     & 5 \\
    \midrule
    MHA-LSTM \cite{messaoud2021attention}& 0.19  & 0.55  & 1.10  & 1.84  & 2.78  \\
    CF-LSTM \cite{xie2021congestion}& 0.18  & 0.42  & 1.07  & 1.72  & 2.44  \\
    EA-Net \cite{cai2021environment} & 0.15  & 0.26  & 0.43  & 0.78  & 1.32  \\
    STDAN \cite{chen2022intention}& 0.19  & 0.27  & 0.48  & 0.91  & 1.66  \\
    DRBP \cite{gao2023dual}& 0.41  & 0.79  & 1.11  & 1.40  & - \\
    WSiP \cite{wang2023wsip}& 0.20  & 0.60  & 1.21  & 2.07  & 3.14  \\
    DACR-AMTP \cite{cong2023dacr} & 0.10  & 0.17  & 0.31  & 0.54  & 1.01  \\
    BAT (25\%) \cite{liao2024bat} & 0.14  & 0.34  &0.65  & 0.89  & 1.27 \\
    BAT \cite{liao2024bat} & {0.08}  & 0.14  & \textbf{0.20}  &  {0.44} & {0.62}\\
     \midrule
    \textbf{NEST} & \textbf{ 0.05 } & \textbf{0.11} &  \textbf{0.20}  & \textbf{ 0.32 } & \textbf{ 0.48 } \\
    \toprule
    \end{tabular}%
    }
  \caption{Experimental results on HighD. Metric: RMSE}
  \label{highdresult}%
\end{table}%

\section{Experiment}\label{Experiment}

\subsection{Experimental Setup}
We conducted experiments on real-world traffic datasets (nuScenes, MoCAD, and HighD) to showcase the superiority of our NEST model over state-of-the-art baselines. We included both quantitative and qualitative analyses, along with inference time comparisons. Each model component's contribution was also evaluated. To ensure consistent and fair comparisons, we followed the established evaluation metrics for each dataset: $\operatorname{minADE}_k$ and $\operatorname{minFDE}_k$ for nuScenes, and Root Mean Squared Error ($\operatorname{RMSE}_k$) for MoCAD and HighD. 

\subsection{Quantitative Results}
The quantitative comparison results for different datasets are detailed in Tables \ref{table:performance_nuscene}, \ref{mocadresult}, and \ref{highdresult}, respectively. Table  \ref{table:performance_nuscene} shows that our NEST model consistently achieves SOTA performance across all metrics on the nuScenes dataset. Notably, it improves the $\text{minADE}_5$ metric by 14.5\% compared to the best existing model \cite{zhang2024seflow}, indicating superior trajectory prediction accuracy. Our model also shows significant enhancements in both the $\text{minADE}_{1}$ and $\text{minFDE}_{1}$ metrics, demonstrating comprehensive improvements in predictive capability. Furthermore, as the value of $k$ decreases, the superiority of our model becomes more evident, showcasing its ability to accurately capture the true intentions of agents.

To validate the effectiveness of our Neuromodulated Small-world Hypergraph in modeling agent interactions, we tested its performance on the MoCAD and HighD datasets without relying on HD maps. As detailed in Table \ref{mocadresult}, our NEST model outperforms all competitors across various prediction horizons in the MoCAD dataset, particularly improving accuracy by 16\% in the 5-second prediction horizon over the previous best model. Although our model slightly trails the best in short-term (2-second) predictions by just 0.01 meter, it still shows strong performance. Table \ref{highdresult} illustrates our model's superiority on the HighD dataset, focused on highway scenarios. Here, the NEST model achieves significant improvements of 27.3\% and 22.6\% over the BAT model \cite{liao2024bat} in the 4-second and 5-second prediction windows, respectively. These results across diverse scenarios highlight the exceptional generalization and predictive capabilities of our NEST model in the real world.

\begin{table}[tbp]
\centering
\resizebox{0.95\linewidth}{!}
{
\begin{tabular}{cc}
\toprule
\text{Model} & Inference Time (ms)  \\
\midrule
P2T \cite{deo2020trajectory} & 116  \\
Trajectron++ \cite{salzmann2020trajectron++} &     38   \\
MultiPath \cite{chai2020multipath} &   87   \\
AgentFormer \cite{yuan2021agentformer} &   107   \\
PGP \cite{deo2022multimodal} & 215  \\
LAformer \cite{liu2024laformer} & 115 \\
VisionTrap \cite{moon2024visiontrap} & 53\\
 \midrule
\textbf{NEST (RTX 3090)} & 11.6  \\
\bottomrule
\end{tabular}}
\caption{Inference time comparison on nuScenes dataset.}
\label{table:performance_inference}
\end{table}

\begin{figure*}[t]
        \centering
	\includegraphics[width=0.98\linewidth]{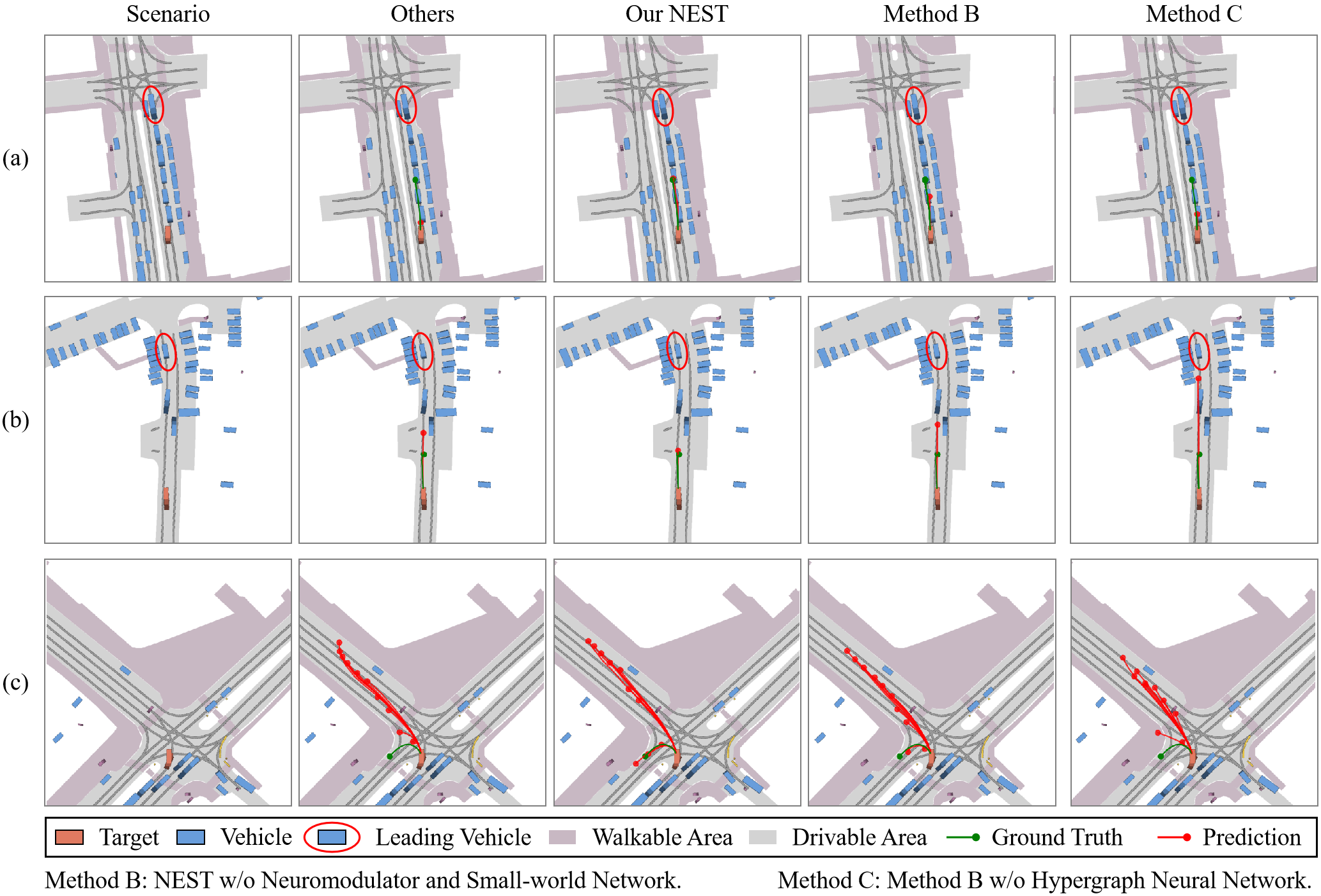}
	\caption{Qualitative comparison of our NEST model with various models. Panels (a) and (b) visualize the most probable prediction by each model, whereas panel (c) visualizes predictions across ten modalities. Panel (a) illustrates the scenario where a distant leading vehicle accelerates. In this scenario, our NEST model maintains a trajectory prediction that aligns closely with the ground truth. In contrast, other models are misled by the stationary state of surrounding vehicles, resulting in slower predictions. Panel (b) depicts a scenario where a distant leading vehicle remains stationary, significantly influencing the expected speed of the target agent. Again, only our NEST model predicts a trajectory that mirrors the ground truth closely, while other models predict overly fast trajectories. Panel (c) distinctly showcases the efficacy of hypergraph-based models (NEST and Method B) in predicting the driver's true intent through their prediction of a complex U-turn maneuver.}
        \label{keshihua}
\end{figure*}

\subsection{Inference Time Comparison}
To demonstrate the efficiency of our proposed NEST model, we conducted a comparative experiment on inference times on nuScenes dataset. Table \ref{table:performance_inference} presents the inference times of various models, including baseline figures sourced from VisionTrap \cite{moon2024visiontrap}, which reported the time required to predict trajectories for 12 agents using a single RTX 3090 Ti GPU. For a fair comparison, we evaluated the NEST model by calculating its average inference time for 12 agents across the dataset. Due to the unavailability of an RTX 3090 Ti GPU, our experiments were performed with an RTX 3090 GPU. As evidenced in Table \ref{table:performance_inference}, the NEST model consistently demonstrates superior inference speed compared to existing baselines even with a slower GPU than RTX 3090 Ti, highlighting its real-time performance.

\subsection{Ablation Studies}
The NEST model is composed of multiple components that collectively yield impressive performance. However, it remains to be seen whether each component significantly contributes to the model's predictive capability. To quantify the impact of each component on the model's performance, we conducted a series of ablation studies. We present detailed ablation variants of the NEST model on the nuScenes dataset in Table \ref{ab2_strategy_nuScenes} and \ref{ab2_nuscenes}. Method F represents the complete NEST model, which achieves the SOTA performance across all metrics, demonstrating the synergistic effect of its components. Method C, which employs a conventional graph neural network to replace hypergraph learning, exhibits the poorest performance across most metrics, highlighting the significance of the Neuromodulated Small-world Hypergraph in accurately learning agent interactions. Method B, which discards the Neuromodulated Small-world Network approach for constructing the interaction hypergraph, exhibits significantly poorer performance across all metrics. Method D, which lacks the context fusion module, also shows degraded performance, indicating the importance of contextual information in driving scenarios for precise trajectory prediction. The results of Method A, and Method E show varying degrees of performance degradation, further confirming the necessity of each component within the NEST model.

\begin{table}[tbp]
  \centering
  \setlength{\tabcolsep}{3mm}
  \resizebox{\linewidth}{!}{
            \begin{tabular}{cccccccc}
                \toprule
                \multirow{2}[4]{*}{Components} & \multicolumn{6}{c}{Ablation Methods} \\
                \cmidrule{2-7}          & A     & B     & C     & D     & E  &F  \\
                \midrule
                Neuromodulator & \ding{56} & \ding{56} & \ding{56} & \ding{52} & \ding{52} & \ding{52} \\
                Small-world Network & \ding{52} & \ding{56} & \ding{56}& \ding{52}  & \ding{52} & \ding{52}  \\
                 Hypergraph Learning& \ding{52} & \ding{52} &  \ding{56} &\ding{52} & \ding{52}  & \ding{52} \\
                 Context Fusion & \ding{52} & \ding{52} & \ding{52} & \ding{56} & \ding{52}  & \ding{52}\\
                 Multi-modal Decoder & \ding{52} & \ding{52} & \ding{52} & \ding{52}& \ding{56}  & \ding{52}\\
                \bottomrule
            \end{tabular}}%
  \caption{Ablation setting of nuScenes.}
  \label{ab2_strategy_nuScenes}%
\end{table}%

\begin{table}[tbp]
  \centering
  \setlength{\tabcolsep}{3mm}
  \resizebox{\linewidth}{!}
  {
            \begin{tabular}{cccccccc}
            \toprule
            \multirow{2}[4]{*}{Metrics} & \multicolumn{6}{c}{Ablation Methods} \\
        \cmidrule{2-7}          & A     & B     & C     & D     & E  &F \\
            \midrule
             $\text{mADE}_5$ & 1.21 & 1.25 & 1.48 & 1.37 & 1.22 & 1.18 \\
             $\text{mADE}_1$ & 2.99 & 3.13 & 3.21 & 3.17 & 3.06 & 2.97 \\
             $\text{mFDE}_1$ & 6.92 & 7.28 & 7.44 & 7.32 & 7.12 & 6.87 \\
            \bottomrule
            \end{tabular}}%
      \caption{Ablation results on nuScenes.}
  \label{ab2_nuscenes}%
\end{table}%

\subsection{Qualitative Results}
In addition to the quantitative analysis, we present the qualitative results from the nuScenes dataset to provide an intuitive understanding of the interaction comprehension capabilities and predictive accuracy of our NEST model. Figure \ref{keshihua} provides a summary of the predictive results across various scenarios for the others model \cite{chen2024q}, ours (NEST model, and its ablated variants—method B and method C). Panels (a) and (b) illustrate the small-world phenomenon in traffic scenarios, where the distant leading vehicle could influence the target agent's behavior. The comparison of various models demonstrates the efficacy of Small-world Networks on interaction modeling within these scenarios. Modeling the inherent uncertainty in driving behavior is crucial for accurate prediction. To demonstrate the NEST model’s capability in modeling uncertainty, we present the multimodal prediction results from various models in panel (c). Panel (c) elucidates the proficiency of our NEST in capturing the intention of drivers.

\section{Conclusion}\label{Conclusion}
In this paper, we introduce the NEST model, an innovative framework designed to address critical challenges in trajectory prediction for autonomous driving. By integrating the strengths of Small-world Networks and hypergraphs, NEST captures both local and long-range interactions among heterogeneous traffic agents in an efficient manner. The Neuromodulator component further enhances the model's adaptability to dynamic traffic conditions, ensuring accurate and reliable predictions. Extensive validation across multiple real-world datasets, including nuScenes, MoCAD, and HighD, consistently demonstrates the NEST model's superior performance and generalization capabilities in various traffic scenarios.

\section{Acknowledgements}\label{Acknowledgements}
This research is supported by Science and Technology Development Fund of Macau SAR (File no. 0021/2022/ITP, 0081/2022/A2, 001/2024/SKL), Shenzhen-Hong Kong-Macau Science and Technology Program Category C (SGDX20230821095159012), State Key Lab of Intelligent Transportation System (2024-B001), Jiangsu Provincial Science and Technology Program (BZ2024055), and University of Macau (SRG2023-00037-IOTSC).

\bibliography{aaai25}

\end{document}